%% file: main.tex
\documentclass[runningheads]{llncs}
\usepackage{graphicx}
\usepackage{comment}
\usepackage{amsmath,amssymb} 
\usepackage{color}
\usepackage{cite}
\usepackage{threeparttable}
\usepackage{comment}
\usepackage{adjustbox}
\usepackage{array,multirow}
\usepackage{colortbl}
\usepackage[table]{xcolor}
\usepackage{makecell}
\usepackage[percent]{overpic}

\usepackage{booktabs}
\usepackage{arydshln}

\newcommand{\pp}{p}
\newcommand{\RR}{\mathbb{R}}
\newcommand{\FF}{\mathbf{F}}
\newcommand{\XX}{\mathbf{X}}
\newcommand{\YY}{\mathbf{Y}}

\begin{document}
\pagestyle{headings}
\mainmatter
\def\ECCVSubNumber{2785}  

\title{Semantic Flow for Fast and Accurate Scene Parsing} 

\titlerunning{ECCV-20 submission ID \ECCVSubNumber} 
\authorrunning{ECCV-20 submission ID \ECCVSubNumber} 
\author{Anonymous ECCV submission}
\institute{Paper ID \ECCVSubNumber}

\author{Xiangtai Li\inst{1} \thanks{The first two authors have equal contribution.} \and
Ansheng You\inst{1} \textsuperscript{*} \and
Zhen Zhu\inst{2} \and
Houlong Zhao\inst{3} \and
Maoke Yang\inst{3} \and
Kuiyuan Yang\inst{3} \and
Shaohua Tan\inst{1} \and
Yunhai Tong\inst{1}
}
\titlerunning{Semantic Flow for Fast and Accurate Scene Parsing} 
\authorrunning{X. Li et al.}
%
\institute{Key Laboratory of Machine Perception, MOE, School of EECS, Peking University \and Huazhong University of Science and Technology \and 
DeepMotion
}
\maketitle

\input{0abstract}
\input{1introduction}

\input{2relatedwork}
\input{3method}

\input{4experiment}
\input{5conclusion}
\input{6sub}

\clearpage
%
%
\bibliographystyle{splncs04}
\bibliography{egbib}
\end{document}

%% file: 0abstract.tex
\begin{abstract}
In this paper, we focus on designing effective method for fast and accurate scene parsing. A common practice to improve the performance is to attain high resolution feature maps with strong semantic representation. Two strategies are widely used---atrous convolutions and feature pyramid fusion, are either computation intensive or ineffective. Inspired by the Optical Flow for motion alignment between adjacent video frames, we propose a Flow Alignment Module (FAM) to learn Semantic Flow between feature maps of adjacent levels, and broadcast high-level features to high resolution features effectively and efficiently. Furthermore, integrating our module to a common feature pyramid structure exhibits superior performance over other real-time methods even on light-weight backbone networks, such as ResNet-18. Extensive experiments are conducted on several challenging datasets, including Cityscapes, PASCAL Context, ADE20K and CamVid. Especially, our network is the first to achieve 80.4\% mIoU on Cityscapes with a frame rate of 26 FPS. The code is available at \url{https://github.com/lxtGH/SFSegNets}.

\keywords{Scene Parsing, Semantic Flow, Flow Alignment Module}
\end{abstract}

%% file: 1introduction.tex
\section{Introduction}
Scene parsing or semantic segmentation is a fundamental vision task which aims to classify each pixel in the images correctly. Two important factors that are highly influential to the performance are: detailed information~\cite{unet} and strong semantics representation~\cite{pspnet,deeplabv3}.
The seminal work of Long~\emph{et. al.}~\cite{fcn} built a deep Fully Convolutional Network (FCN), which is mainly composed from convolutional layers, in order to carve strong semantic representation. However, detailed object boundary information, which is also crucial to the performance, is usually missing due to the use of the down-sampling layers. To alleviate this problem, state-of-the-art methods~\cite{pspnet,psanet,DAnet,nvidia_seg_video} apply atrous convolutions~\cite{dilation} at the last several stages of their networks to yield feature maps with strong semantic representation while at the same time maintaining the high resolution. 

\begin{figure}
	\centering
	\includegraphics[width=0.5\linewidth]{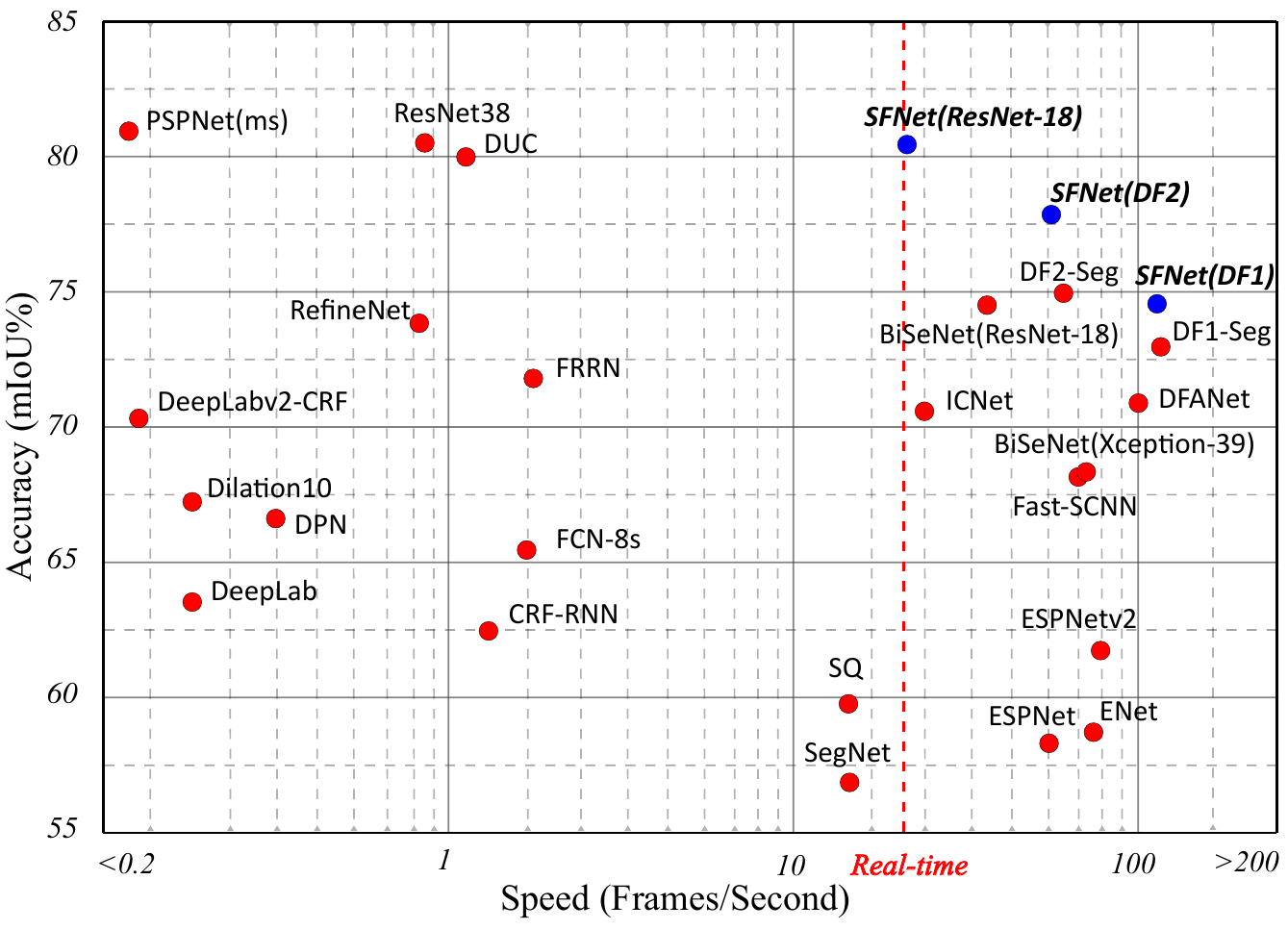}
	\caption{Inference speed versus mIoU performance on test set of Cityscapes. Previous models are marked as red points, and our models are shown in blue points which achieve the best speed/accuracy trade-off. Note that our method with ResNet-18 as backbone even achieves comparable accuracy with all accurate models at much faster speed.}
	\label{fig:teaser}
\end{figure}

Nevertheless, doing so inevitably requires intensive extra computation since the feature maps in the last several layers can reach up to 64 times bigger than those in FCNs.
Given that the FCN using ResNet-18~\cite{resnet} as the backbone network has a frame rate of 57.2 FPS for a $1024\times 2048$ image, after applying atrous convolutions~\cite{dilation} to the network as done in \cite{pspnet,psanet}, the modified network only has a frame rate of 8.7 FPS. Moreover, under a single GTX 1080Ti GPU with no other ongoing programs, the previous state-of-the-art model PSPNet~\cite{pspnet} has a frame rate of only 1.6 FPS for $1024 \times 2048$ input images. As a consequence, this is very problematic to many advanced real-world applications, such as self-driving cars and robots navigation, which desperately demand real-time online data processing. 

In order to not only maintain detailed resolution information but also get features that exhibit strong semantic representation, another direction is to build FPN-like~\cite{fpn,PanopticFPN,unet} models which leverage the lateral path to fuse feature maps in a top-down manner. In this way, the deep features of the last several layers strengthen the shallow features with high resolution and therefore, the refined features are possible to satisfy the above two factors and beneficial to the accuracy improvement. However, 
the accuracy of these methods~\cite{unet,segnet} is still unsatisfactory when compared to those networks who hold large feature maps in the last several stages. 
We suspect the low accuracy problem arises from the ineffective propagation of semantics from deep layers to shallow layers. 

To mitigate this issue, we propose to learn the \textbf{Semantic Flow} between two network layers of different resolutions. 
The concept of Semantic Flow is inspired from optical flow, which is widely used in video processing task~\cite{DFF} to represent the pattern of apparent motion of objects, surfaces, and edges in a visual scene caused by relative motion. In a flash of inspiration, we feel the relationship between two feature maps of arbitrary resolutions from the same image can also be represented with the ``motion'' of every pixel from one feature map to the other one. 
In this case, once precise Semantic Flow is obtained, the network is able to propagate semantic features with minimal information loss. 
It should be noted that Semantic Flow is apparently different from optical flow, since Semantic Flow takes feature maps from different levels as input and assesses the discrepancy within them to find a suitable flow field that will give dynamic indication about how to align these two feature maps effectively. 

Based on the concept of Semantic Flow, we design a novel network module called Flow Alignment Module~(FAM) to utilize Semantic Flow in the scene parsing task. Feature maps after FAM are embodied with both rich semantics and abundant spatial information.
Because FAM can effectively transmit the semantic information from deep layers to shallow layers through very simple operations, it shows superior efficacy in both improving the accuracy and keeping superior efficiency. Moreover, FAM is end-to-end trainable, and can be plugged into any backbone networks to improve the results with a minor computational overhead. For simplicity, we call the networks that all incorporate FAM but have different backbones as \textbf{SFNet(backbone)}.
As depicted in Figure~\ref{fig:teaser}, SFNet with different backbone networks outperforms other competitors by a large margin under the same speed. In particular, our method adopting ResNet-18 as backbone achieves \textbf{80.4\%} mIoU on the Cityscapes test server with a frame rate of \textbf{26 FPS}. When adopting DF2~\cite{DF-seg-net} as backbone, our method achieves 77.8\% mIoU with 61 FPS and 74.5\% mIoU with 121 FPS when equipped with the DF1 backbone. Moreover, when using deeper backbone networks, such as ResNet-101, SFNet achieves better results(81.8 \%mIoU) than the previous state-of-the-art model DANet~\cite{DAnet}(81.5 \%mIoU), and only requires \textbf{33\%} computation of DANet during the inference. Besides, the consistent superior efficacy of SFNet across various datasets also clearly demonstrates its broad applicability.

To conclude, our main contributions are three-fold:
\begin{itemize}
\item We introduce the concept of Semantic Flow in the field of scene parsing and propose a novel flow-based align module (FAM) to learn the Semantic Flow between feature maps of adjacent levels and broadcast high-level features to high resolution features more effectively and efficiently.

\item We insert FAMs into the feature pyramid framework and build a feature pyramid aligned network called SFNet for fast and accurate scene parsing.

\item Detailed experiments and analysis indicate the efficacy of our proposed module in both improving the accuracy and keeping light-weight. We achieve state-of-the-art results on Cityscapes, Pascal Context, Camvid datasets and a considerable gain on ADE20K.

\end{itemize}

%% file: 2relatedwork.tex
\section{Related Work}
For scene parsing, there are mainly two paradigms for high-resolution semantic map prediction. One paradigm tries to keep both spatial and semantic information along the main network pathway, while the other paradigm distributes spatial and semantic information to different parts in a network, then merges them back via different strategies.

The first paradigm mostly relies on some network operations to retain high-resolution feature maps in the latter network stages. Many state-of-the-art accurate methods~\cite{pspnet,DAnet,nvidia_seg_video} follow this paradigm to design sophisticated head networks to capture contextual information. PSPNet~\cite{pspnet} proposes to leverage pyramid pooling module (PPM) to model multi-scale contexts, whilst DeepLab series~\cite{deeplabv2,deeplabv3,deeplabv3p,denseaspp} uses astrous spatial pyramid pooling (ASPP). In~\cite{DAnet,AdaptivePyramid_seg,ocnet,emanet,annet,ccnet,DMF_seg}, non-local operator~\cite{non_local} and self-attention mechanism~\cite{transformer} are adopted to harvest pixel-wise context from the whole image. Meanwhile, several works~\cite{graph_cnn,beyond_grids,spg_net,DGM_net,zhangli_dgcn} use graph convolutional neural networks to propagate information over the image by projecting features into an interaction space.

The second paradigm contains several state-of-the-art fast methods, where high-level semantics are represented by low-resolution feature maps. A common strategy is to fuse multi-level feature maps for high-resolution spatiality and strong semantics~\cite{fcn,segnet,upernet,unet,xiangtl_gff}. ICNet~\cite{ICnet} uses multi-scale images as input and a cascade network to be more efficient. DFANet~\cite{dfanet} utilizes a light-weight backbone to speed up its network and proposes a cross-level feature aggregation to boost accuracy, while SwiftNet~\cite{swiftnet} uses lateral connections as the cost-effective solution to restore the prediction resolution while maintaining the speed. To further speed up, low-resolution images are used as input for high-level semantics~\cite{ICnet,gum_bmvc} which reduce features into low resolution and then upsample them back by a large factor. The direct consequence of using a large upsample factor is performance degradation, especially for small objects and object boundaries. Guided upsampling~\cite{gum_bmvc} is related to our method, where the semantic map is upsampled back to the input image size guided by the feature map from an early layer. However, this guidance is still insufficient for some cases due to the information gap between the semantics and resolution. In contrast, our method aligns feature maps from adjacent levels and further enhances the feature maps using a feature pyramid framework towards both high resolution and strong semantics, consequently resulting in the state-of-the-art performance considering the trade-off between high accuracy and fast speed. 

There is another set of works focusing on designing light-weight backbone networks to achieve real-time performances. ESPNets~\cite{ESPNet,ESPNetv2} save computation by decomposing standard convolution into point-wise convolution and spatial pyramid of dilated convolutions. BiSeNet~\cite{bisenet} introduces spatial path and semantic path to reduce computation. Recently, several methods~\cite{fast_cell_search_seg,custom_search_seg,DF-seg-net} use AutoML techniques to search efficient architectures for scene parsing. Our method is complementary to some of these works, which further boosts their accuracy. 
Since our proposed semantic flow is inspired by optical flow~\cite{FlowNet}, which is used in video semantic segmentation, we also discuss several works in video semantic segmentation. 
For accurate results, temporal information is exceedingly exploited by using optical flow. 
Gadde~\emph{et. al.}~\cite{Netwarp} warps internal feature maps and Nilsson~\emph{et. al.}~\cite{GRFP_video} warps final semantic maps from nearby frame predictions to the current map. To pursue faster speed, optical flow is used to bypass the low-level feature computation of some frames by warping features from their preceding frames~\cite{DFF,Lowlatency_net}. Our work is different from theirs by propagating information hierarchically in another dimension, which is orthogonal to the temporal propagation for videos. 

%% file: 3method.tex
\vspace{-2ex} 
\section{Method}

\begin{figure*}[!t]
	\centering
	\includegraphics[width=1.0\linewidth]{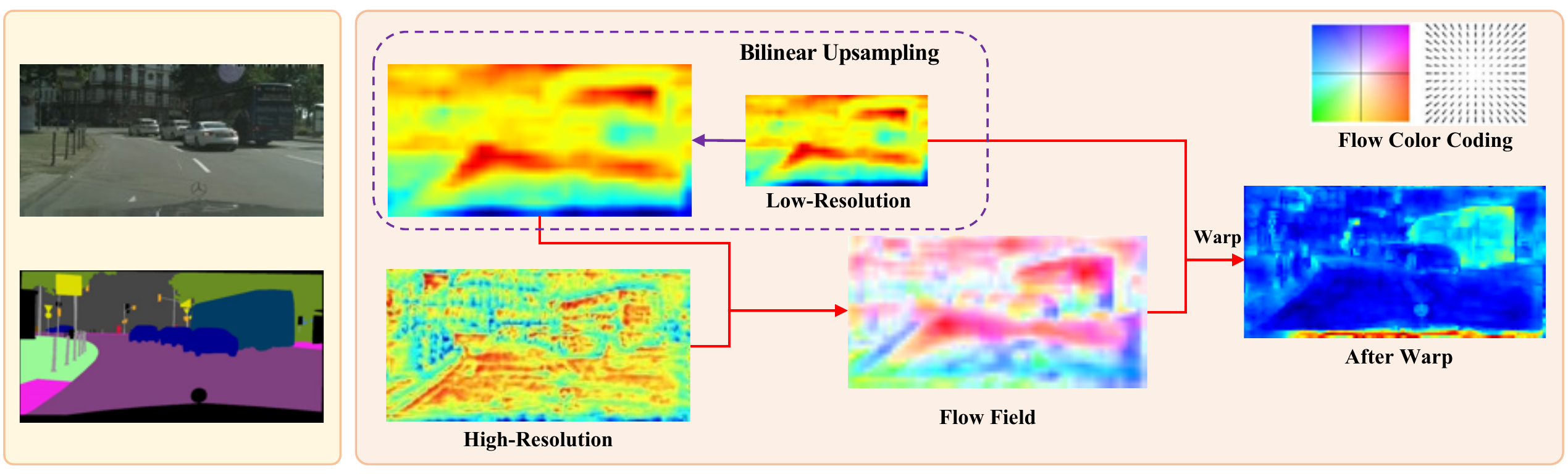}
	\caption{Visualization of feature maps and semantic flow field in FAM. Feature maps are visualized by averaging along the channel dimension. Larger values are denoted by hot colors and vice versa. We use the color code proposed in~\cite{flowvis} to visualize the Semantic Flow field. The orientation and magnitude of flow vectors are represented by hue and saturation respectively. 
	}
	\label{fig:issue}
\end{figure*}

In this section, we will first give some preliminary knowledge about scene parsing and introduce the misalignment problem therein. Then, we propose the Flow Alignment Module (FAM) to resolve the misalignment issue by learning Semantic Flow and warping top-layer feature maps accordingly. Finally, we present the whole network architecture equipped with FAMs based on the FPN framework~\cite{fpn} for fast and accurate scene parsing.
\vspace{-1ex} 
\subsection{Preliminary}
The task of scene parsing is to map an RGB image $\XX \in \RR^{H\times W \times 3}$ to a semantic map $\YY \in \RR^{H\times W \times C}$ with the same spatial resolution $H\times W$, where $C$ is the number of predefined semantic categories. Following the setting of FPN~\cite{fpn}, the input image $\XX$ is firstly mapped to a set of feature maps $\{\FF_l\}_{l=2,...,5}$ from each network stage, where $\FF_l \in \RR^{H_l \times W_l \times C_l}$ is a $C_l$-dimensional feature map defined on a spatial grid $\Omega_l$ with size of $H_l \times W_l, H_l = \frac{H}{2^l}, W_l = \frac{W}{2^l}$.
The coarsest feature map $\FF_5$ comes from the deepest layer with strongest semantics. FCN-32s directly predicts upon $\FF_5$ and achieves over-smoothed results without fine details. However, some improvements can be achieved by fusing predictions from lower levels~\cite{fcn}. FPN takes a step further to gradually fuse high-level feature maps with low-level feature maps in a top-down pathway through $2\times$ bi-linear upsampling, which was originally proposed for object detection~\cite{fpn} and recently introduced for scene parsing~\cite{upernet,PanopticFPN}. 
The whole FPN framework highly relies on upsampling operator to upsample the spatially smaller but semantically stronger feature map to be larger in spatial size. However, the bilinear upsampling recovers the resolution of downsampled feature maps by interpolating a set of uniformly sampled positions (i.e., it can only handle one kind of fixed and predefined misalignment), while the misalignment between feature maps caused by a residual connection, repeated downsampling and upsampling, is far more complex. Therefore, position correspondence between feature maps needs to be explicitly and dynamically established to resolve their actual misalignment.

\begin{figure}[!t]
	\centering
	\includegraphics[width=1.0\linewidth]{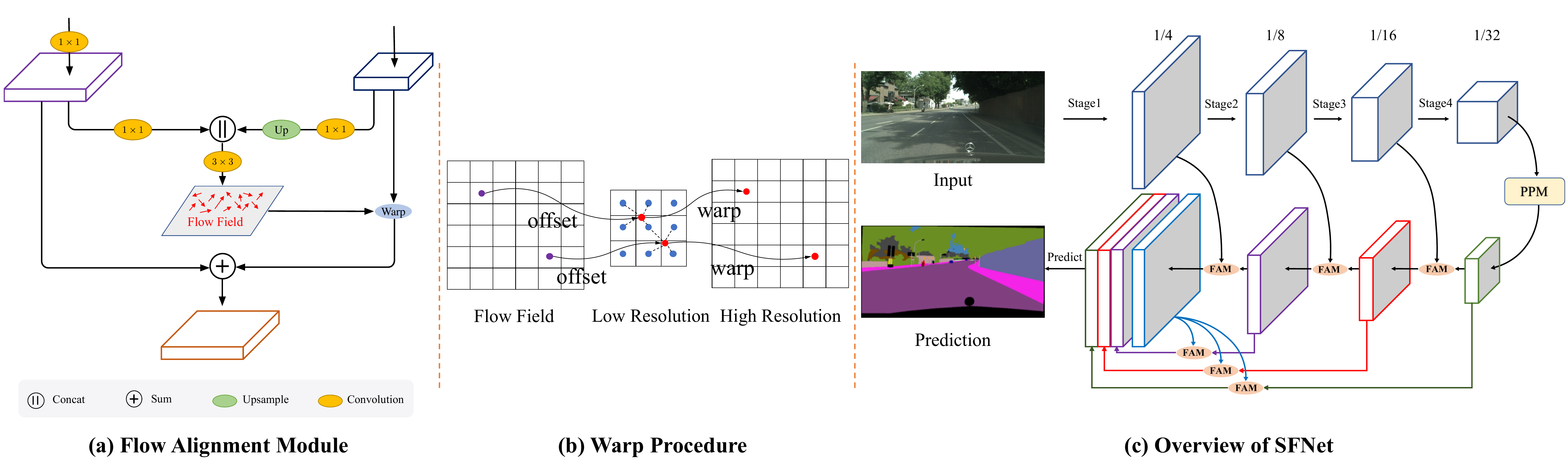}
	\caption{(a) The details of Flow Alignment Module. We combine the transformed high-resolution feature map and low-resolution feature map to generate the semantic flow field, which is utilized to warp the low-resolution feature map to high-resolution feature map. (b) Warp procedure of Flow Alignment Module. The value of the high-resolution feature map is the bilinear interpolation of the neighboring pixels in low-resolution feature map, where the neighborhoods are defined according learned semantic flow field. (c) Overview of our proposed SFNet. ResNet-18 backbone with four stages is used for exemplar illustration. FAM: Flow Alignment Module. PPM: Pyramid Pooling Module~\cite{pspnet}. Best view it in color and zoom in.
	}
	\label{fig:overview}
\end{figure}

\vspace{-1ex} 
\subsection{Flow Alignment Module}

\noindent \textbf{Design Motivation.}~For more flexible and dynamic alignment, we thoroughly investigate the idea of optical flow, which is very effective and flexible to align two adjacent video frame features in the video processing task~\cite{brox2004high,DFF}. The idea of optical flow motivates us to design a \emph{flow-based alignment module} (\textbf{FAM}) to align feature maps of two adjacent levels by predicting a flow field inside the network. We define such flow field as \emph{Semantic Flow}, which is generated between different levels in a feature pyramid. For efficiency, while designing our network, we adopt an efficient backbone network---FlowNet-S~\cite{FlowNet}.

\noindent \textbf{Module Details.}
FAM is built within the FPN framework, where feature map of each level is compressed into the same channel depth through two 1$\times$1 convolution layers before entering the next level. 
Given two adjacent feature maps $\FF_{l}$ and $\FF_{l-1}$ with the same channel number, we up-sample $\FF_{l}$ to the same size as $\FF_{l-1}$ via a bi-linear interpolation layer. Then, we concatenate them together and take the concatenated feature map as input for a sub-network that contains two convolutional layers with the kernel size of $3\times 3$. The output of the sub-network is the prediction of the semantic flow field $\Delta_{l-1} \in \RR^{H_{l-1} \times W_{l-1} \times 2}$. Mathematically, the aforementioned steps can be written as:
\begin{equation}
    \Delta_{l-1} = \text{conv}_l(\text{cat}(\FF_{l}, \FF_{l-1})),
\end{equation}
where $\text{cat}(\cdot)$ represents the concatenation operation and $\text{conv}_l(\cdot)$ is the $3\times 3$ convolutional layer.
Since our network adopts strided convolutions, which could lead to very low resolution, for most cases, the respective field of the 3$ \times $3 convolution $\text{conv}_l$ is sufficient to cover most large objects of that feature map. Note that, we discard the correlation layer proposed in FlowNet-C~\cite{FlowNet}, where positional correspondence is calculated explicitly. Because there exists a huge semantic gap between higher-level layer and lower-level layer, explicit correspondence calculation on such features is difficult and tends to fail for offset prediction. Moreover, adopting such a correlation layer introduces heavy computation cost, which violates our goal for the network to be fast and accurate.

After having computed $\Delta_{l-1}$, each position $\pp_{l-1}$ on the spatial grid $\Omega_{l-1}$ is then mapped to a point $\pp_{l}$ on the upper level $l$ via a simple addition operation. Since there exists a resolution gap between features and flow field shown in Fig~\ref{fig:overview}(b), the warped grid and its offset should be halved as Eq~\ref{mapping},
\begin{equation}
    \pp_{l} = \frac{\pp_{l-1}+\Delta_{l-1}(\pp_{l-1})}{2}. 
    \label{mapping}
\end{equation}
We then use the differentiable bi-linear sampling mechanism proposed in the spatial transformer networks~\cite{STN}, which linearly interpolates the values of the 4-neighbors (top-left, top-right, bottom-left, and bottom-right) of $\pp_{l}$ to approximate the final output of the FAM, denoted by $\widetilde\FF_l(\pp_{l-1})$. Mathematically,
\begin{equation}
    \widetilde\FF_l(\pp_{l-1}) = \FF_l(\pp_{l}) = \sum_{\pp \in \mathcal{N}(\pp_{l})} w_\pp \FF_{l}(\pp),
    \label{interpolate}
\end{equation}
where 
$\mathcal{N}(\pp_{l})$ represents neighbors of the warped points $\pp_l$ in $\FF_l$ and $w_\pp $ denotes the bi-linear kernel weights estimated by the distance of warped grid. This warping procedure may look similar to the convolution operation of the deformable kernels in deformable convolution network~(DCN)~\cite{deformable}. However, our method has a lot of noticeable difference from DCN. First, our predicted offset field incorporates both higher-level and lower-level features to \emph{align the positions} between high-level and low-level feature maps, while the offset field of DCN moves the positions of the kernels according to the predicted location offsets in order to \emph{possess larger and more adaptive respective fields}. Second, our module focuses on aligning features while DCN works more like an attention mechanism that attends to the salient parts of the objects. More detailed comparison can be found in the experiment part.

On the whole, the proposed FAM module is light-weight and end-to-end trainable because it only contains one 3$\times$3 convolution layer and one parameter-free warping operation in total. Besides these merits, it can be plugged into networks multiple times with only a minor extra computation cost overhead. Figure~\ref{fig:overview}(a) gives the detailed settings of the proposed module while Figure~\ref{fig:overview}(b) shows the warping process. Figure~\ref{fig:issue} visualizes feature maps of two adjacent levels, their learned semantic flow and the finally warped feature map. As shown in Figure~\ref{fig:issue}, the warped feature is more structurally neat than normal bi-linear upsampled feature and leads to more consistent representation of objects, such as the bus and car.


\vspace{-1ex} 
\subsection{Network Architectures}
Figure~\ref{fig:overview}(c) illustrates the whole network architecture, which contains a bottom-up pathway as the encoder and a top-down pathway as the decoder. While the encoder has a backbone network offering feature representations of different levels, the decoder can be seen as a FPN equipped with several FAMs.

\noindent \textbf{Encoder Part.} We choose standard networks pre-trained on ImageNet~\cite{imagenet} for image classification as our backbone network by removing the last fully connected layer. Specifically, ResNet series~\cite{resnet}, ShuffleNet v2~\cite{shufflenetv2}  and DF series~\cite{DF-seg-net} are used and compared in our experiments. All backbones have $4$ stages with residual blocks, and each stage has a convolutional layer with stride 2 in the first place to downsample the feature map chasing for both computational efficiency and larger receptive fields. We additionally adopt the Pyramid Pooling Module (PPM)~\cite{pspnet} for its superior power to capture contextual information. In our setting, the output of PPM shares the same resolution as that of the last residual module. In this situation, we treat PPM and the last residual module together as the last stage for the upcoming FPN. Other modules like ASPP~\cite{deeplabv3} can also be plugged into our network, which are also experimentally ablated in Sec.~\ref{sec:ablation}.

\noindent \textbf{Aligned FPN Decoder} takes feature maps from the encoder and uses the aligned feature pyramid for final scene parsing. By replacing normal bi-linear up-sampling with FAM in the top-down pathway of FPN~\cite{fpn}, $\{\FF_l\}_{l=2}^4$ is refined to $\{\widetilde{\FF}_l\}_{l=2}^4$, where top-level feature maps are aligned and fused into their bottom levels via element-wise addition and $l$ represents the range of feature pyramid level. For scene parsing, $\{\widetilde{\FF}_l\}_{l=2}^4 \cup \{\FF_5\}$ are up-sampled to the same resolution (\emph{i.e.}, 1/4 of input image) and concatenated together for prediction. Considering there are still misalignments during the previous step, we also replace these up-sampling operations with the proposed FAM.

\noindent \textbf{Cascaded Deeply Supervised Learning.} We use deeply supervised loss~\cite{pspnet} to supervise intermediate outputs of the decoder for easier optimization. In addition, following~\cite{bisenet}, online hard example mining~\cite{ohem} is also used by only training on the $10\%$ hardest pixels sorted by cross-entropy loss.

%% file: 4experiment.tex
\section{Experiments}
We first carry out experiments on the Cityscapes~\cite{Cityscapes} dataset, which is comprised of a large set of high-resolution $(2048 \times 1024)$ images in street scenes. This dataset has 5,000 images with high quality pixel-wise annotations for 19 classes, which is further divided into 2975, 500, and 1525 images for training, validation and testing. To be noted, coarse data are not used in this work. Besides, more experiments on Pascal Context~\cite{VOC}, ADE20K~\cite{ADE20K} and CamVid~\cite{CamVid} are summarised to further prove the generality of our method. 

\subsection{Experiments on Cityscapes}

\noindent
\textbf{Implementation details:} We use PyTorch~\cite{pytorch} framework to carry out following experiments. All networks are trained with the same setting, where stochastic gradient descent (SGD) with batch size of 16 is used as optimizer, with momentum of 0.9 and weight decay of 5e-4. All models are trained for 50K iterations with an initial learning rate of 0.01. As a common practice, the ``poly'' learning rate policy is adopted to decay the initial learning rate by multiplying $(1 -\frac{\text{iter}}{\text{total}\_\text{iter}})^{0.9}$ during training. Data augmentation contains random horizontal flip, random resizing with scale range of $[0.75, ~2.0]$, and random cropping with crop size of $1024 \times 1024$. During inference, we use the whole picture as input to report performance unless explicitly mentioned. For quantitative evaluation, mean of class-wise intersection-over-union (mIoU) is used for accurate comparison, and number of float-point operations (FLOPs) and frames per second (FPS) are adopted for speed comparison.

\noindent
\textbf{Comparison with baseline methods: } Table~\ref{tab:city_ablation_architecture}(a) reports the  comparison results against baselines on the validation set of Cityscapes~\cite{Cityscapes}, where ResNet-18~\cite{resnet} serves as the backbone. Comparing with the naive FCN, dilated FCN improves mIoU by 1.1\%.  By appending the FPN decoder to the naive FCN, we get 74.8\% mIoU by an improvement of 3.2\%. By replacing bilinear upsampling with the proposed FAM, mIoU is boosted to 77.2\%, which improves the naive FCN and FPN decoder by 5.7\% and 2.4\% respectively. Finally, we append PPM (Pyramid Pooling Module)~\cite{pspnet} to capture global contextual information, which achieves the best mIoU of 78.7 \% together with FAM. Meanwhile, FAM is complementary to PPM by observing FAM improves PPM from 76.6\% to 78.7\%.

\begin{table}[!t]
	\centering
		\begin{minipage}{\textwidth}
	    \begin{minipage}{\dimexpr.45 \linewidth}		
			\centering
			\resizebox{0.75\textwidth}{!}{%
				\begin{tabular}{l c c c}
				\toprule[0.2em]
				Method &Stride & mIoU ($\%$) & $\Delta a (\%)$ \\
				\toprule[0.2em]
					FCN& 32 & 71.5  & -   \\
					Dilated FCN&8 & 72.6 & 1.1 $\uparrow$ \\
				\midrule
					+FPN &32 & 74.8 & 3.3 $\uparrow$ \\
					+FAM &32 & 77.2 & 5.7 $\uparrow$ \\ 
					+FPN + PPM &32 & 76.6 & 5.1 $\uparrow$ \\
					+FAM + PPM&32  & \textbf{78.7} & 7.2 $\uparrow$ \\
				\bottomrule[0.1em]
			    \end{tabular}
			}\par
			{(a) Ablation study on baseline model.}
		\end{minipage}
		\begin{minipage}{\dimexpr.45 \linewidth}
			\centering
			\resizebox{0.75\textwidth}{!}{%
			\begin{tabular}{c c c c l l}
			\toprule[0.2em]
			Method & $\FF_3$ & $\FF_4$ & $\FF_5$ & mIoU(\%) & $\Delta a(\%) $ \\  
			\toprule[0.2em]
			FPN+PPM &  &  &  & 76.6 & -  \\ 
			& \checkmark & & & 76.9 & 0.3 $\uparrow$ \\
			& & \checkmark & & 77.0 & 0.4 $\uparrow$ \\
			& & & \checkmark & 77.5 & 0.9 $\uparrow$ \\
			\midrule
			\midrule
			& &\checkmark &\checkmark & 77.8 & 1.2 $\uparrow$\\
			&\checkmark &\checkmark &\checkmark &78.3& 1.7 $\uparrow$\\
			\hline
		\end{tabular}
			}\par
	
			{ (b) Ablation study on insertion position.}
			
		\end{minipage}
		
	\end{minipage}
	
	\begin{minipage}{\textwidth}
		\centering
		\begin{minipage}{\dimexpr.45 \linewidth}
		    \centering
			\resizebox{0.75\textwidth}{!}{%
				\begin{tabular}{l l l l }
				\toprule[0.2em]
				Method & mIoU(\%) & $\Delta a(\%) $ & \#GFLOPs \\  
				\toprule[0.2em]
				FAM & 76.4  & -  & - \\
				+PPM~\cite{pspnet} & 78.3 & 1.9$\uparrow$ & 123.5 \\
				+NL~\cite{non_local} & 76.8 & 0.4$\uparrow$ & 148.0 \\
				+ASPP~\cite{deeplabv3} & 77.6 & 1.2$\uparrow$ & 138.6 \\
				+DenseASPP~\cite{denseaspp} & 77.5 & 1.1$\uparrow$ & 141.5 \\
				\hline
			\end{tabular}
			}\par
			{(c) Ablation study on context module.}
			
		\end{minipage}	
		\begin{minipage}{\dimexpr.45 \linewidth}
		    \centering
			\resizebox{0.65\textwidth}{!}{%
		    \begin{tabular}{l c c c}
				\toprule[0.2em]
				Backbone & mIoU(\%) & $\Delta a(\%) $ & \#GFLOPs \\  
				\toprule[0.2em]
				ResNet-50~\cite{resnet}& 76.8  & - & 332.6 \\
				w/ FAM & 79.2 & 2.4 $\uparrow$& 337.1 \\ 
				ResNet-101~\cite{resnet}& 77.6 & - & 412.7 \\
				w/ FAM  & 79.8 & 2.2$\uparrow$ & 417.5 \\
				\midrule
				ShuffleNetv2~\cite{shufflenetv2} & 69.8 & - & 17.8 \\
				w/ FAM  & 72.1 & 2.3 $\uparrow$& 18.1 \\
				DF1~\cite{DF-seg-net} & 72.1 & - & 18.6 \\
				w/ FAM  & 74.3 & 2.2 $\uparrow$ & 18.7 \\
				DF2~\cite{DF-seg-net} & 73.2 & - & 48.2\\
				w/ FAM  & 75.8 & 2.6 $\uparrow$ & 48.5 \\
				\bottomrule[0.1em]
			\end{tabular}
			}\par
			{(d) Ablation on study on various backbones.}
		\end{minipage}
	\end{minipage}

	\caption{Experiments results on network design using Cityscapes validation set.}
	\label{tab:city_ablation_architecture}
\end{table}


\noindent
\textbf{Positions to insert FAM:} We insert FAM to different stage positions in the FPN decoder and report the results as Table~\ref{tab:city_ablation_architecture}(b). From the first three rows, FAM improves all stages and gets the greatest improvement at the last stage, which demonstrate that misalignment exists in all stages on FPN and is more severe in coarse layers. This is consistent with the fact that coarse layers containing stronger semantics but with lower resolution, and can greatly boost segmentation performance when they are appropriately upsampled to high resolution. The best result is achieved by adding FAM to all stages in the last row. Note that, for fast speed, we adopt FAMs only in the adjacent feature pyramids.

\noindent
\textbf{Ablation study on network architecture design: } \label{sec:ablation}
Considering current state-of-the-art contextual modules are used as heads on dilated backbone networks~\cite{deeplabv3, Co-Occurrent, DAnet, pspnet, psanet, denseaspp}, we further try different contextual heads in our methods where coarse feature map is used for contextual modeling. Table~\ref{tab:city_ablation_architecture}(c) reports the comparison results, where PPM~\cite{pspnet} delivers the best result, while more recently proposed methods such as Non-Local based heads~\cite{non_local} perform worse. Therefore, we choose PPM as our contextual head considering its better performance with lower computational cost. We further carry out experiments with different backbone networks including both deep and light-weight networks, where FPN decoder with PPM head is used as a strong baseline in Table~\ref{tab:city_ablation_architecture}(d). For heavy networks, we choose ResNet-50 and ResNet-101~\cite{resnet} as representation. For light-weight networks, ShuffleNetv2~\cite{shufflenetv2} and DF1/DF2~\cite{DF-seg-net} are employed. FAM significantly achieves better mIoU on all backbones with slightly extra computational cost.

\begin{table}[!t]
	\centering
	\begin{minipage}{\textwidth}
	    \begin{minipage}{\dimexpr.45 \linewidth}		
			\centering
			\resizebox{0.65\textwidth}{!}{%
				\begin{tabular}{l c}
				\toprule[0.2em]
				Method  & mIoU ($\%$)  \\
				\toprule[0.2em]
					bilinear upsampling & 78.3 \\
					deconvolution & 77.9 \\ 
					nearest neighbor & 78.2 \\
 				\bottomrule[0.1em]
			    \end{tabular}
			}\par
			{(a) Ablation study on Upsampling operation in FAM.}
		\end{minipage}
		\begin{minipage}{\dimexpr.45 \linewidth}
			\centering
			\resizebox{0.50\textwidth}{!}{%
			\begin{tabular}{l c c}
				\toprule[0.2em]
				Method  & mIoU ($\%$) & Gflops\\
				\toprule[0.2em]
				    $k=1$ & 77.8 & 120.4 \\
				    $k=3$ & 78.3 & 123.5 \\
				    $k=5$ & 78.1 & 131.6 \\
                    $k=7$ & 78.0 & 140.5 \\
				\bottomrule[0.1em]
			    \end{tabular}
			}\par
			{ (b) Ablation study on kernel size $k$ in FAM where 3 FAMs are involved.}
			
		\end{minipage}
		
	\end{minipage}
	
	\begin{minipage}{\textwidth}
		\centering
		\begin{minipage}{\dimexpr.45 \linewidth}
		    \centering
			\resizebox{0.65\textwidth}{!}{%
			\begin{tabular}{l c c c}
				\toprule[0.2em]
				Method  & mIoU ($\%$) & $\Delta a(\%) $ \\
				\toprule[0.2em]
					FPN +PPM  & 76.6  & -   \\
				\midrule
				    correlation~\cite{FlowNet} & 77.2 & 0.6 $\uparrow$ \\
					Ours  & 77.5 & 0.9 $\uparrow$ \\
				\bottomrule[0.1em]
			    \end{tabular}
			}\par
			{(c) Ablation with FlowNet-C~\cite{FlowNet} in FAM.}
			
		\end{minipage}	
		\begin{minipage}{\dimexpr.45 \linewidth}
		    \centering
			\resizebox{0.60\textwidth}{!}{%
			 \begin{tabular}{c c c c l l}
			\toprule[0.2em]
			Method & $\FF_3$ & $\FF_4$ & $\FF_5$ & mIoU(\%) & $\Delta a(\%) $ \\  
			\toprule[0.2em]
			FPN +PPM & -  & - & - &  76.6 & - \\
			DCN & &  &\checkmark & 76.9 & 0.3 $\uparrow$ \\
			Ours & &  &\checkmark & 77.5 & 0.9 $\uparrow$ \\
			\midrule
			\midrule
		    DCN	&\checkmark & \checkmark & \checkmark & 77.2 & 0.6 $\uparrow$\\
			Ours &\checkmark &\checkmark & \checkmark & 78.3 & 1.7 $\uparrow$\\
			\hline
		\end{tabular}
			}\par
			{(d) Comparison with DCN~\cite{deformable}.}
		\end{minipage}
	\end{minipage}

	\caption{Experiments results on FAM design using Cityscapes validation set.}
	\label{tab:city_ablation_FAM}
\end{table}


\noindent
\textbf{Ablation study on FAM design:} We first explore the effect of upsampling in FAM in Table~\ref{tab:city_ablation_FAM}(a). Replacing the bilinear upsampling with deconvolution and nearest neighbor upsampling achieves 77.9 mIoU and 78.2 mIoU, respectively, which are similar to the 78.3 mIoU achieved by bilinear upsampling. We also try the various kernel size in Table~\ref{tab:city_ablation_FAM}(b). Larger kernel size of $5\times5$ is also tried which results in a similar (78.2) but introduces more computation cost. In Table~\ref{tab:city_ablation_FAM}(c), replacing FlowNet-S with correlation in FlowNet-C also leads to slightly worse results (77.2) but increases the inference time. The results show that it is enough to use lightweight FlowNet-S for aligning feature maps in FPN. In Table~\ref{tab:city_ablation_FAM}(d), we compare our results with DCN~\cite{deformable}. We apply DCN on the concatenated feature map of bilinear upsampled feature map and the feature map of next level. We first insert one DCN in higher layers $\FF_{5}$ where our FAM is better than it. After applying DCN to all layers, the performance gap is much larger. This denotes our method can also align low level edges for better boundaries and edges in lower layers, which will be shown in visualization part.

\begin{figure*}[!t]
	\centering
	\includegraphics[width=0.75\linewidth]{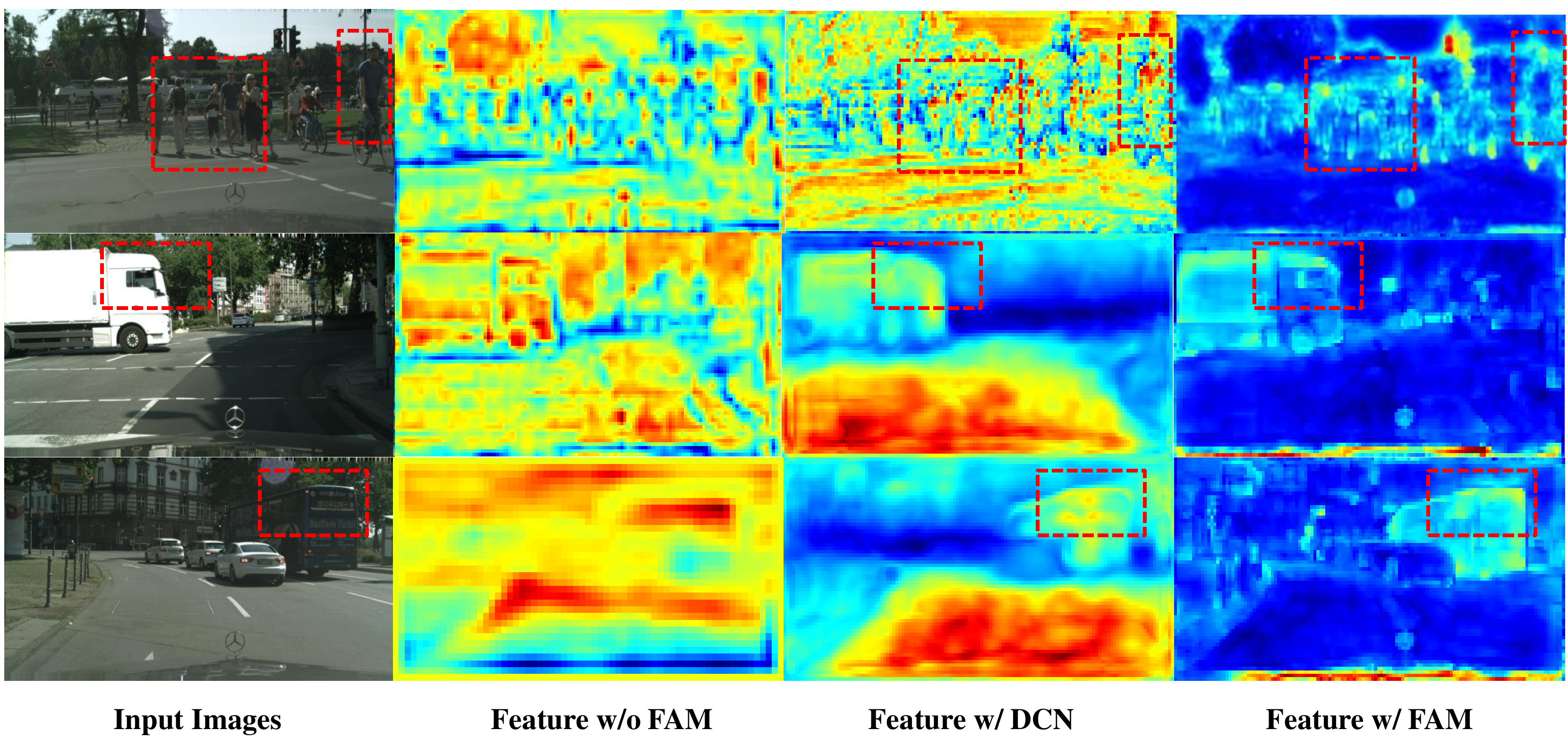}
	\caption{
	Visualization of the aligned feature. Compared with DCN, our module outputs more structural feature representation. 
	}
	\label{fig:vis_align_fea}
\end{figure*}

\noindent
\textbf{Aligned feature representation:} In this part, we give more visualization on aligned feature representation as shown in Figure~\ref{fig:vis_align_fea}. We visualize the upsampled feature in the final stage of ResNet-18. It shows that compared with DCN~\cite{deformable}, our FAM feature is more structural and has much more precise objects boundaries which is consistent with the results in Table~\ref{tab:city_ablation_FAM}(d). That indicates FAM is \textbf{not} an attention effect on feature similar to DCN, but actually aligns feature towards more precise shape as compared in red boxes. 

\begin{figure*}[!t]
	\centering
	\includegraphics[width=1.0\linewidth]{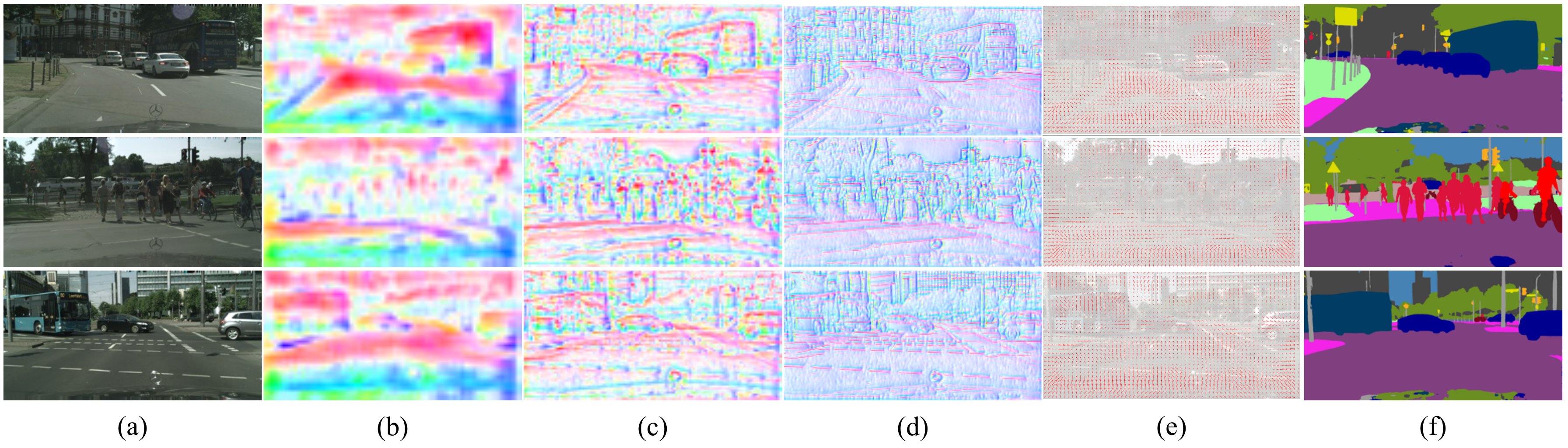}
	\caption{Visualization of the learned semantic flow fields. Column (a) lists three exemplary images. Column (b)-(d) show the semantic flow of the three FAMs in an ascending order of resolution during the decoding process, following the same color coding of Figure~\ref{fig:issue}. Column (e) is the arrowhead visualization of flow fields in column (d). Column (f) contains the segmentation results.
	}
	\label{fig:vis_flowfield}
\end{figure*}

\noindent
\textbf{Visualization of Semantic Flow: } 
Figure~\ref{fig:vis_flowfield} visualizes semantic flow from FAM in different stages. Similar with optical flow, semantic flow is visualized by color coding and is bilinearly interpolated to image size for quick overview. Besides, vector fields are also visualized for detailed inspection. From the visualization, we observe that semantic flow tends to diffuse out from some positions inside objects, where these positions are generally near object centers and have better receptive fields to activate top-level features with pure, strong semantics. Top-level features at these positions are then propagated to appropriate high-resolution positions following the guidance of semantic flow. In addition, semantic flows also have coarse-to-fine trends from top level to bottom level, which phenomenon is consistent with the fact that semantic flows gradually describe offsets between gradually smaller patterns.

\begin{figure}[!t]
	\centering
	\includegraphics[width=1.0\linewidth]{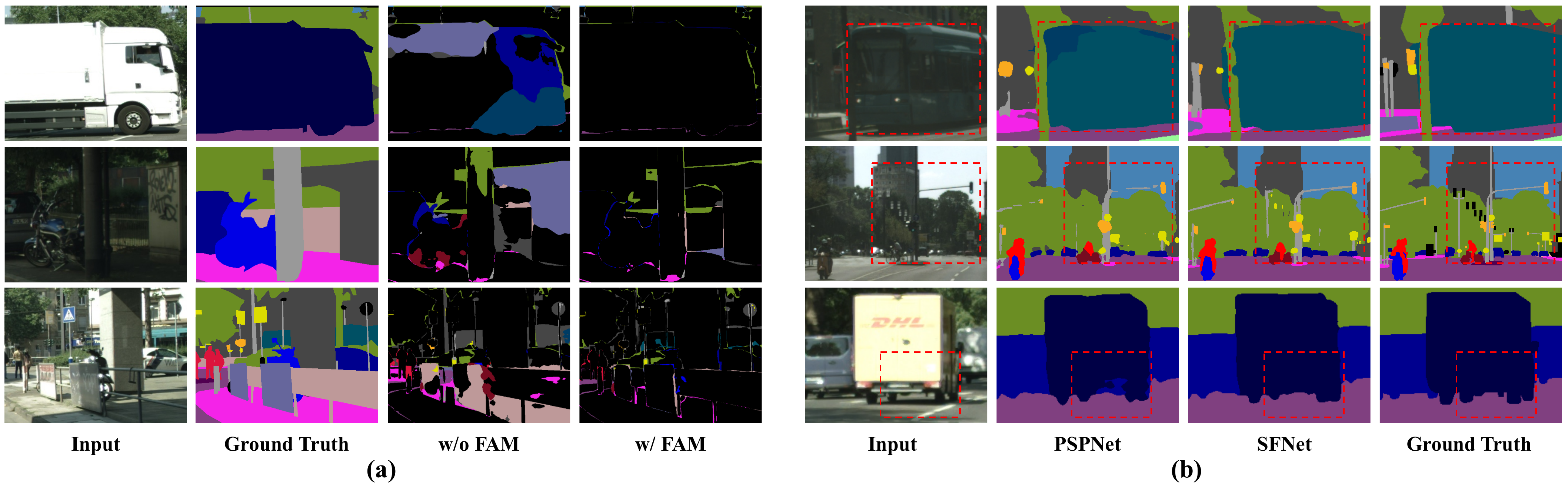}
	\caption{(a), Qualitative comparison in terms of errors in predictions, where correctly predicted pixels are shown as black background while wrongly predicted pixels are colored with their groundtruth label color codes.
	(b), Scene parsing results comparison against PSPNet~\cite{pspnet}, where significantly improved regions are marked with red dashed boxes. Our method performs better on both small scale and large scale objects.
	}
	\label{fig:error_map}
\end{figure}

\noindent
\textbf{Visual Improvement analysis:} Figure~\ref{fig:error_map}(a) visualizes the prediction errors by both methods, where FAM considerably resolves ambiguities inside large objects (e.g., truck) and produces more precise boundaries for small and thin objects (e.g., poles, edges of wall). Figure\ref{fig:error_map} (b) shows our model can better handle the small objects with shaper boundaries than dilated PSPNet due to the alignment on lower layers.

\begin{table}[!t]\setlength{\tabcolsep}{6pt}
	\centering
	\begin{threeparttable}
		\scalebox{0.65}{
			\begin{tabular}{l c c c c}
				\toprule[0.2em]
				Method  &   InputSize & mIoU ($\%$) & \#FPS & \#Params \\
				\toprule[0.2em]
				ENet~\cite{ENnet} &  $640\times360$ &  58.3  & 60 & 0.4M \\
				ESPNet~\cite{ESPNet} & $512 \times 1024$ & 60.3 & 132 & 0.4M \\
				ESPNetv2~\cite{ESPNetv2} &  $512 \times 1024$ & 62.1 & 80 & 0.8M \\
				ERFNet~\cite{ERFNet} & $512 \times 1024 $& 69.7 & 41.9 & - \\
				BiSeNet(ResNet-18)~\cite{bisenet} &  $768 \times 1536$ &  $74.6$ & 43 & 12.9M  \\
				BiSeNet(Xception-39)~\cite{bisenet} &  $768 \times 1536$ &  $68.4$ & 72 & 5.8M  \\ 
				ICNet~\cite{ICnet} & $1024 \times 2048$ &  69.5 & 34 & 26.5M \\
				DF1-Seg~\cite{DF-seg-net} & $1024 \times 2048$ & 73.0 & 80 & 8.55M  \\
				DF2-Seg~\cite{DF-seg-net} & $1024 \times 2048$ & 74.8 & 55 & 8.55M  \\
				SwiftNet~\cite{swiftnet} & $1024 \times 2048$  & 75.5 & 39.9 & 11.80M \\
				SwiftNet-ens~\cite{swiftnet} & $1024 \times 2048$  & 76.5 & 18.4 & 24.7M \\
				DFANet~\cite{dfanet} & $1024 \times 1024$ & 71.3 & 100 & 7.8M \\
				CellNet~\cite{custom_search_seg} & $768 \times 1536$ & 70.5  & 108 & -\\
				\midrule
				SFNet(DF1) & $1024 \times 2048$ & $\bf{74.5}$ & 74/\underline{121} & 9.03M \\
				SFNet(DF2) & $1024 \times 2048$ & $\bf{77.8}$ & 53/\underline{61} & 10.53M \\
				SFNet(ResNet-18) & $1024 \times 2048$ & $\bf{78.9}$ & 18/\underline{26} & 12.87M \\
				SFNet(ResNet-18)\textdagger & $1024 \times 2048$ & $\bf{80.4}$ & 18/\underline{26} & 12.87M \\
				\bottomrule[0.1em]
			\end{tabular}
		}
		\begin{tablenotes}
			 \item {\scriptsize \textdagger Mapillary dataset used for pretraining.
			 }
		\end{tablenotes}
		\caption{Comparison on Cityscapes {\it test} set with state-of-the-art real-time models.  For fair comparison, input size is also considered, and all models use single scale inference. }
		\label{table:cityscapes_sota_speed_acc}
	\end{threeparttable}
\end{table}

\noindent
\textbf{Comparison with real-time models:} All compared methods are evaluated by single-scale inference and input sizes are also listed for fair comparison. Our speed is tested on one GTX 1080Ti GPU with full image resolution $1024 \times 2048$ as input, and we report speed of two versions, i.e., without and with TensorRT acceleration. As shown in Table~\ref{table:cityscapes_sota_speed_acc}, our method based on DF1 achieves a more accurate result(74.5\%) than all methods faster than it. With DF2, our method outperforms all previous methods while running at 60 FPS. With ResNet-18 as backbone, our method achieves 78.9\% mIoU and even reaches performance of accurate models which will be discussed in the next experiment. By additionally using Mapillary~\cite{mapillary} dataset for pretraining, our ResNet-18 based model achieves 26 FPS with 80.4\% mIoU, which sets the new state-of-the-art record on accuracy and speed trade-off on Cityscapes benchmark. More detailed information are in the supplementary file.

\begin{table}[!t]\setlength{\tabcolsep}{6pt}
	\centering
		\begin{threeparttable}
		\scalebox{0.65}{
			\begin{tabular}{l c c c c}
				\toprule[0.2em]
				Method  & Backbone & mIoU ($\%$) & \#Params & \#GFLOPs\tnote{\textdagger} \\
				\toprule[0.2em]
				SAC~\cite{sac}  &  ResNet-101  &  $78.1$ & - & - \\
				DepthSeg~\cite{depthseg} &   ResNet-101  &  $78.2$ & -  & - \\ 
				PSPNet~\cite{pspnet}  &   ResNet-101  &  $78.4$ & $65.7$M & $1065.4$ \\ 
				BiSeNet~\cite{bisenet} & ResNe-18 & 77.7 & 12.3M & 82.2 \\
				BiSeNet~\cite{bisenet} &   ResNet-101  &  $78.9$ & $51.0$M & $219.1$ \\ 
				DFN~\cite{dfn} &  ResNet-101  &  $79.3$ & $90.7$M & 1121.0 \\ 
				PSANet~\cite{psanet} &   ResNet-101  &  $80.1$ & 85.6M & 1182.6 \\ 
				DenseASPP~\cite{denseaspp}  &  DenseNet-161  &  $80.6$ & $35.7$M & $632.9$\\
				SPGNet~\cite{SPGNet} & 2$\times$ResNet-50 & 81.1 & - & -\\
				ANNet~\cite{annet} & ResNet-101 & 81.3 & 63.0M & 1089.8 \\ 
				CCNet~\cite{ccnet} & ResNet-101 & 81.4 & 66.5M & 1153.9 \\
				DANet~\cite{DAnet} & ResNet-101 & 81.5 & 66.6M & 1298.8\\ 
				\midrule
				SFNet &  ResNet-18 & $\bf{79.5}$ & \textbf{$\mathbf{12.87}$M} & $\mathbf{123.5}$\\
				SFNet &  ResNet-101 & $\bf{81.8}$ & \textbf{$\mathbf{50.32}$M} & $\mathbf{417.5}$ \\
				\bottomrule[0.1em]			\end{tabular}
		}
		\begin{tablenotes}
			\item[\textdagger] {\scriptsize \#GFLOPs calculation adopts  $1024 \times 1024$ image as input.}
		\end{tablenotes}
		\caption{Comparison  on Cityscapes {\it test} set with state-of-the-art accurate models. For better accuracy, all models use multi-scale inference. }
		\label{table:cityscapes_sota_acc}
	\end{threeparttable}
\end{table}

\noindent
\textbf{Comparison with accurate models: } 
State-of-the-art accurate models~\cite{DAnet,pspnet,denseaspp,nvidia_seg_video} perform multi-scale and horizontal flip inference to achieve better results on the Cityscapes test server. For fair comparison, we also report multi-scale with flip testing results following previous methods~\cite{pspnet,DAnet}. Model parameters and computation FLOPs are also listed for comparison. Table~\ref{table:cityscapes_sota_acc} summarizes the results, where our models achieve state-of-the-art accuracy while costs much less computation. In particular, our method based on ResNet-18 is 1.1\% mIoU higher than PSPNet~\cite{pspnet} while only requiring \textbf{11\%} of its computation. Our ResNet-101 based model achieves better results than DAnet~\cite{DAnet} by 0.3\% mIoU and only requires \textbf{30\%} of its computation.

\subsection{Experiment on More Datasets}
 We also perform more experiments on other three data-sets including Pascal Context~\cite{pcontext-data}, ADE20K~\cite{ADE20K} and CamVid~\cite{CamVid} to further prove the effectiveness of our method. More detailed setting can be found in the supplemental file.

\begin{table}[!t]
	\centering
	\begin{minipage}{\textwidth}
		\centering
		\begin{minipage}{\dimexpr.45 \linewidth}
		    \centering
			\resizebox{0.85\textwidth}{!}{%
			\begin{tabular}{l c c c }
			\toprule[0.2em]
			Method & Backbone & mIoU (\%) & \#GFLOPs\tnote{\textdagger} \\
			\toprule[0.2em]
			Ding~\textit{et al.}~\cite{ding2018context} & ResNet-101  & 51.6 & - \\
			EncNet~\cite{context_encoding} &ResNet-50   & 49.2 & - \\
			EncNet~\cite{context_encoding} &ResNet-101   &51.7 & - \\
			DANet~\cite{DAnet}&ResNet-50  & 50.1  &  186.4 \\
			DANet~\cite{DAnet}&ResNet-101  & 52.6  & 257.1 \\
			ANNet~\cite{annet} & ResNet-101 & 52.8 & 243.8 \\
			BAFPNet~\cite{BAFPNet} & ResNet-101 & 53.6 & - \\
			EMANet~\cite{emanet} & ResNet-101 & 53.1 & 209.3 \\
			\hline 
			w/o FAM & ResNet-50 & 49.0 & 74.5 \\
			SFNet & ResNet-50   & \textbf{50.7}(1.7 $\uparrow$) & 75.4 \\
			w/o FAM & ResNet-101 & 51.1 & 92.7\\
			SFNet & ResNet-101  & \textbf{53.8}(2.7 $\uparrow$)& 93.6 \\
			\bottomrule[0.1em]
			\end{tabular}
			}\par
			{(a) Results on Pascal Context. Evaluated on 60 classes. }
			
		\end{minipage}	
		\begin{minipage}{\dimexpr.45 \linewidth}
		    \centering
			\resizebox{0.85\textwidth}{!}{%
		    \begin{tabular}{l l l l}
				\toprule[0.2em]
				Method & Backbone & mIoU (\%) & \#GFLOPs\tnote{\textdagger}  \\
				\toprule[0.2em]
				PSPNet~\cite{pspnet} & ResNet-50 & 42.78 & 167.6 \\
				PSPNet~\cite{pspnet} & ResNet-101 &  43.29 & 238.4 \\ 
				PSANet~\cite{psanet} & ResNet-101 &  43.77 & 264.9 \\ 
				EncNet~\cite{context_encoding} & ResNet-101 & 44.65 & - \\
				CFNet~\cite{Co-Occurrent} & ResNet101 & 44.82 & -\\
				\midrule
				w/o FAM & ResNet-50 & 41.12 & 74.8 \\
				SFNet & ResNet-50  & 42.81(1.69 $\uparrow$)  & 75.7 \\
				w/o FAM & ResNet-101 & 43.08 & 93.1 \\
				SFNet & ResNet-101  & 44.67(1.59 $\uparrow$) & 94.0 \\
				\bottomrule[0.1em]
			\end{tabular}
			}\par
			{(b) Results on ADE20K.}
		\end{minipage}
	\end{minipage}
	\caption{Experiments results on Pascal Context and ADE20k(Multi scale inference).  \#GFLOPs calculation adopts $480 \times 480$ image as input.}
	\label{tab:pacal_context_ade_20k}
\end{table}

\noindent
\textbf{PASCAL Context: } The results are illustrated as Table~\ref{tab:pacal_context_ade_20k}(a), our method outperforms corresponding baselines by 1.7\% mIoU and 2.6\% mIoU with ResNet-50 and ResNet-101 as backbones respectively. In addition, our method on both ResNet-50 and ResNet-101 outperforms their existing counterparts by large margins with significantly lower computational cost.

\noindent
\textbf{ADE20K: } is a challenging scene parsing dataset. Images in this dataset are from different scenes with more scale variations. Table~\ref{tab:pacal_context_ade_20k}(b) reports the performance comparisons, our method improves the baselines by 1.69\% mIoU and 1.59\% mIoU respectively, and outperforms previous state-of-the-art methods~\cite{pspnet,psanet} with much less computation.

\noindent
\textbf{CamVid: } is another road scene dataset. This dataset involves 367 training images, 101 validation images and 233 testing images with resolution of $960 \times 720$. We apply our method with different light-weight backbones on this dataset and report comparison results in Table~\ref{table:camvid_res}. With DF2 as backbone, FAM improves its baseline by 3.2\% mIoU. Our method based on ResNet-18 performs best with 73.8\% mIoU while running at 35.5 FPS.

\begin{table}[!t]\setlength{\tabcolsep}{6pt}
	\centering
	\begin{threeparttable}
		\scalebox{0.70}{
			\begin{tabular}{l  l l l}
				\toprule[0.2em]
				Method  &  Backbone & mIoU (\%) & FPS \\
				\toprule[0.2em]
			    ICNet~\cite{ICnet}  & ResNet-50 & 67.1 & 34.5 \\
			    BiSegNet~\cite{bisenet} & Xception-39 & 65.6 & - \\
			    BiSegNet~\cite{bisenet} & ResNet-18 & 68.7 & - \\
			    DFANet A~\cite{dfanet} & - & 64.7 & 120 \\
			    DFANet B~\cite{dfanet} & - & 59.3 & 160 \\
				\midrule
                w/o FAM &DF2 & 67.2 & 139.8 \\
				SFNet & DF2 & \textbf{70.4} (3.2 $\uparrow$) & 134.1 \\
				SFNet & ResNet-18 & \textbf{73.8} & 35.5 \\
				\bottomrule[0.1em]
			\end{tabular}
		}
		\caption{Accuracy and Speed comparison with previous state-of-the-art real-time models on CamVid~\cite{CamVid} test set where the input size is $960 \times 720$ with single scale inference.
		}
		\label{table:camvid_res}
	\end{threeparttable}
\end{table}

%% file: 5conclusion.tex
\section{Conclusion}
In this paper, we devise to use the learned \textbf{Semantic Flow} to align multi-level feature maps generated by a feature pyramid to the task of scene parsing. With the proposed flow alignment module, high-level features are well fused into low-level feature maps with high resolution. By discarding atrous convolutions to reduce computation overhead and employing the flow alignment module to enrich the semantic representation of low-level features, our network achieves the best trade-off between semantic segmentation accuracy and running time efficiency. Experiments on multiple challenging datasets illustrate the efficacy of our method. 

%% file: 6sub.tex
\section{Supplemental Parts}

Our supplemental material contains two parts. One is the more details on Cityscapes datasets and the other is the detailed setting on other datasets. We will opensource the our codebase.

\section{Supplemental Experiments on Cityscapes}

\noindent
\textbf{Detailed improvement on baseline models:}
Table~\ref{tab:cityscapes_results_detail_val} compares the detailed results of each category on the validation set, where ResNet-101 is used as backbone, and FPN decoder with PPM head serves as the baseline. Our method improves almost all categories, especially for 'truck' with more than 19\% mIoU improvement.

\begin{figure*}[!t]
	\centering
	\includegraphics[width=1.0\linewidth]{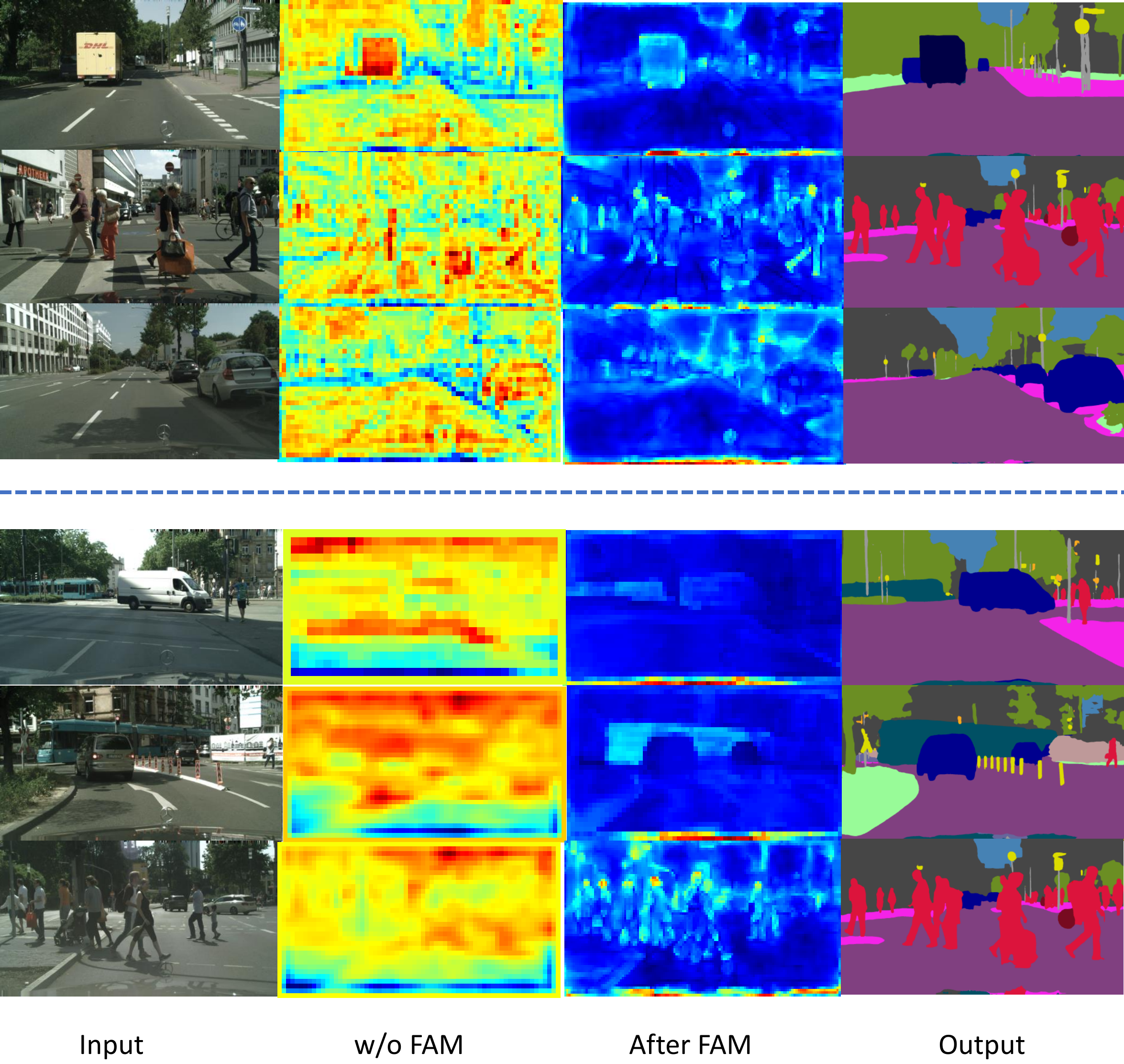}
	\caption{More visualization of the aligned feature representation. The figures below the blue line are the outputs of last stage of FAM while the figures above the blues are the outputs of next to the last stage FAM with more fine details. Best view it on screen.
	}
	\label{fig:more_vis_fea}
\end{figure*}

\vspace{1ex}
\noindent
\textbf{More structured feature visualization on FAM:} We give more structured feature visualization
in Figure~\ref{fig:more_vis_fea}. We visualize more FAM outputs with two different location: last stages(below the blue line) and next to the last stage(above the blue line). For both cases, our module aligns the features into more structured representations with more clear shape and accurate boundaries.

\begin{figure*}[!t]
	\centering
	\includegraphics[width=1.0\linewidth]{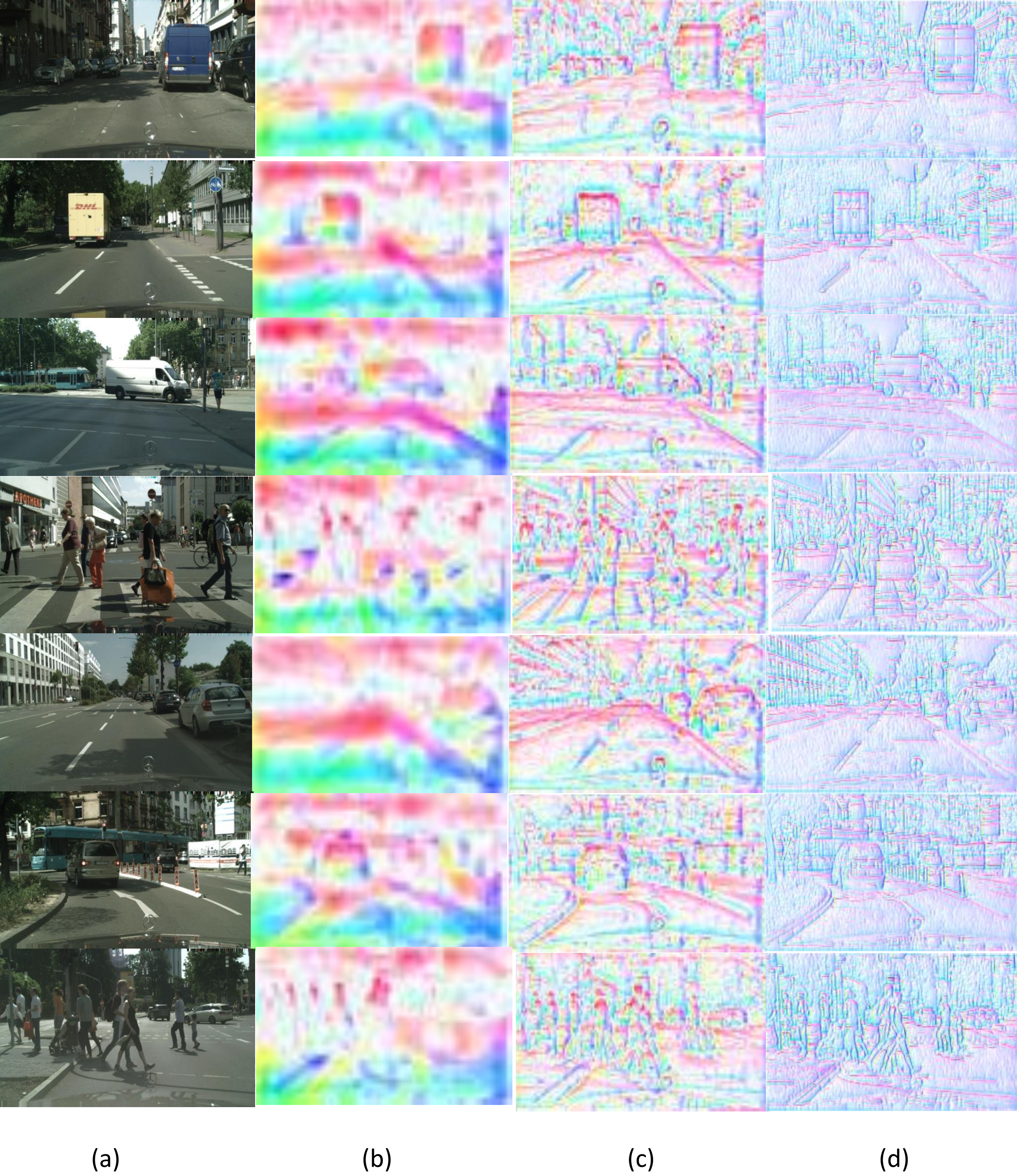}
	\caption{More visualization of the learned semantic flow fields. Column (a) lists input images. Column (b)-(d) show the semantic flow of the three FAMs in an ascending order of resolution during the decoding process. Best view it on screen.
	}
	\label{fig:more_vis_flow}
\end{figure*}

\vspace{1ex}
\noindent
\textbf{More Visualization of Learned Flow:} We also give more learned semantic flow  Visualization in Figure~\ref{fig:more_vis_flow}.

\vspace{1ex}
\noindent
\textbf{More training details using Mapillary Vistas~\cite{mapillary}:}
Mapillary Vistas is a large-scale dataset captured at street scenes, which contains 18K/2K/5K images for training, validation and testing, respectively. The dataset is similar to Cityscapes. Due to the larger variance of image resolutions than Cityscapes, we resize longer side to 2048 before data augmentation. To verify the performance improvement of SFNet by using more training data, we first pre-train SFNet on Mapillary Vistas for 50,000 iterations by using both train and val dataset, then finetune the model on Cityscapes for 50,000 iterations using Cityscapes fine-annotated data with the same setting in the paper before the submission to the test server.

\vspace{1ex}
\noindent
\textbf{Detailed setting about TensorRT}
The testing environment is TensorRT 6.0.1 with CUDA 10.1 on a single GTX 1080Ti GPU. In addition, we re-implement grid sampling operator by CUDA to be used together with TensorRT. The operator is provided by PyTorch and used in warping operation in the Flow Alignment Module. Also, we also test the our speed on 1080-Ti using pytorch-library~\cite{pytorch} where we report average time of inferencing 100 images.

\vspace{1ex}
\noindent
\textbf{Detailed results and settings on Cityscapes compared with accurate models:} We give more detailed results in Table~\ref{tab:cityscapes_results_detail_fine} for the state-of-the-art model comparison. For the fair comparison, we adopt multi-scale inference with 7 scales {0.5,0.75, 1, 1.25, 1.5, 1.75, 2.0} with flip operation. As shown in Table~\ref{tab:cityscapes_results_detail_fine}, our SFNet achi
eves the best performance with less GFlops which has been calculated in the main paper.

\begin{table*}[t!]
\centering 
\small
\addtolength{\tabcolsep}{0pt}
\resizebox{\textwidth}{!}{
			\begin{tabular}{ l | c c c c c c c c c c c c c c c c c c c | c}
		   	\hline
			Method & road & swalk & build & wall & fence & pole & tlight & sign & veg. & terrain & sky & person & rider & car & truck & bus & train & mbike & bike & mIoU \\
			\hline
			BaseLine & 98.1 & 84.9 & 92.6 & 54.8 & 62.2 & 66.0 & 72.8 & 80.8 & 92.4 & 60.6 & 94.8 & 83.1 & 66.0 & 94.9 & 65.9 & 83.9 & 70.5 & 66.0 & 78.9 & 77.6 \\ 
			\hline
			w/ FAM & 98.3 & 85.9 & 93.2 & 62.2 & 67.2 & 67.3 & 73.2 & 81.1 & 92.8 & 60.5 & 95.6 & 83.2 & 65.0 & 95.7 & 84.1 & 89.6 & 75.1 & 67.7 & 78.8 & 79.8 \\
			\hline
		\end{tabular}
		}
\vspace{2mm}
\caption{\small{
		Quantitative per-category comparison results on Cityscapes validation set, where ResNet-101 backbone with the FPN decoder and PPM head serves as the strong baseline. Sliding window crop with horizontal flip is used for testing. Obviously, FAM boosts the performance of almost all the categories.}
}
\label{tab:cityscapes_results_detail_val}
\end{table*}

\begin{table*}[t!]
\centering 
\small
\addtolength{\tabcolsep}{0pt}
\resizebox{\textwidth}{!}{
		\begin{tabular}{ l | c c c c c c c c c c c c c c c c c c c | c}
		    \hline
			Method & road & swalk & build & wall & fence & pole & tlight & sign & veg. & terrain & sky & person & rider & car & truck & bus & train & mbike & bike & mIoU \\
			\hline
			ResNet38~\cite{resnet38} & 98.5 & 85.7 & 93.0 & 55.5 & 59.1 & 67.1 & 74.8 & 78.7 & 93.7 & 72.6 & 95.5 & 86.6 & 69.2 & 95.7 & 64.5 & 78.8 & 74.1 & 69.0 & 76.7 & 78.4 \\
			PSPNet~\cite{pspnet} & 98.6 & 86.2 & 92.9 & 50.8 & 58.8 & 64.0 & 75.6 & 79.0 & 93.4 & 72.3 & 95.4 & 86.5 & 71.3 & 95.9 & 68.2 & 79.5 & 73.8 & 69.5 & 77.2 & 78.4\\
			AAF~\cite{aaf} & 98.5 & 85.6 & 93.0 & 53.8 & 58.9 & 65.9 & 75.0 & 78.4 & 93.7 &
			72.4 & 95.6 & 86.4 & 70.5 & 95.9 & 73.9 & 82.7 & 76.9 & 68.7 & 76.4 & 79.1 \\
			SegModel~\cite{segmodel} & 98.6 & 86.4 & 92.8 & 52.4 & 59.7 & 59.6 & 72.5 & 78.3 & 93.3 & 72.8 & 95.5 & 85.4 & 70.1 & 95.6 & 75.4 & 84.1 & 75.1 & 68.7 & 75.0 & 78.5 \\
			DFN~\cite{dfn} & - & - & - & - & - & - & - & - & - & - & - & - & - & - & - & - & - & - & - & 79.3 \\
			BiSeNet~\cite{bisenet} & - & - & - & - & - & - & - & - & - & - & - & - & - & - & - & - & - & - & - & 78.9 \\
			DenseASPP~\cite{denseaspp}  & 98.7 & 87.1 & 93.4 & 60.7 & 62.7 & 65.6 & 74.6 & 78.5 & 93.6 & 72.5 & 95.4 & 86.2 & 71.9 & 96.0 & {78.0} & {90.3} & 80.7 & 69.7 & 76.8 & 80.6 \\
			BFPNet~\cite{BAFPNet} & 98.7 & 87.1 & 93.5 & 59.8 & {63.4} & 68.9 & 76.8 & 80.9 & 93.7 & 72.8 & 95.5 & 87.0 & 72.1 & 96.0 & 77.6 & 89.0 & 86.9 & 69.2 & 77.6 & 81.4 \\ 
			DANet~\cite{DAnet} & 98.6 & 87.1 & 93.5 & 56.1 & 63.3 & 69.7 & 77.3 & 81.3 & 93.9 & 72.9 & 95.7 & 87.3 & 72.9 & 96.2 & 76.8 & 89.4 & 86.5 & {72.2} & 78.2 & 81.5 \\
			\hline
		SFNet & {98.8} & {87.1} & {93.6} & {63.2} & 62.7 & {68.4} & {75.6} & {80.3} & {93.8} & {71.0} & {95.7} & {87.7} & {73.2} & {96.5} & 75.9 & 92.3 & {89.5} & 71.4 & {78.0} & \textbf{81.8} \\
		\hline
		\end{tabular}
}
\vspace{2mm}
\caption{\small{
		Per-category results on Cityscapes test set. Note that all the models are trained with only fine annotated data. Our method achieves \textbf{81.8\%} mIoU with \textbf{much less} GFlops.}
}
\label{tab:cityscapes_results_detail_fine}
\end{table*}

\section{Detailed Experiment Settings on Other Datasets:}

\vspace{2ex}
\noindent
\textbf{PASCAL Context:} provides detailed semantic labels for whole scenes, and contains 4998 images for training and 5105 images for validation. We train the network for 120 epochs with batch size 16, crop size 512 with initial learning rate 1e-3. For evaluation, we perform multi-scale testing with horizontal flip operation.

\vspace{2ex}
\noindent
\textbf{ADE20k:} is a more challenging scene parsing dataset annotated with 150 classes, and it contains 20K/2K images for training and validation. It has the various objects in the scene. We train the network for 120 epochs with batch size  16,  crop  size  512  and  initial learning rate 1e-2. For final testing, we perform multi-scale testing with horizontal flip operation.

\vspace{2ex}
\noindent
\textbf{CamVid:} is a road scene image segmentation dataset, which provides pixel-wise annotations for 11 semantic categories. There are 367 training images, 101 validation images and 233 testing images. We train the model with 120 epochs and our crop size is set to 640 and learning rate is 1e-3. The batch size is set to 16. For the final testing, we perform the single scale test for the fair comparison.